\documentclass[11pt]{article}

\usepackage[margin=1in]{geometry}
\usepackage[T1]{fontenc}
\usepackage[utf8]{inputenc}
\usepackage{lmodern}
\usepackage{microtype}
\usepackage{amsmath,amssymb}
\usepackage{booktabs}
\usepackage{array}
\usepackage{multirow}
\usepackage{graphicx}
\usepackage{xcolor}
\usepackage{hyperref}
\usepackage[nameinlink,noabbrev]{cleveref}

\hypersetup{
  colorlinks=true,
  linkcolor=blue!50!black,
  citecolor=blue!50!black,
  urlcolor=blue!50!black
}

\title{As X, Do Y: How Persona and Task Combine in Instruction-Tuned LLMs}
\author{Eric Xu \\ Independent Researcher \\ \texttt{youxu@wustl.edu}}
\date{}

\begin{document}
\maketitle

\begin{abstract}
Role prompts of the form \emph{``As X, do Y''} combine a persona specification $X$ with a task specification $Y$ at inference time. We characterize where, in the residual stream of an instruction-tuned language model, this combination is causally simple enough to be approximated additively. For each persona-task pair we compare four prompts -- baseline, persona-only, task-only, and persona-plus-task -- and form a pure persona effect $\Delta_X$, a pure task effect $\Delta_Y$, and an interaction term $\mathrm{Inter} = \Delta_{XY} - \Delta_X - \Delta_Y$.

Near answer formation -- the last prompt position together with the first two generated tokens -- $\Delta_{XY}$ is well approximated by $\Delta_X + \Delta_Y$, and substituting the additive prediction for the clean residual yields downstream output within a small KL of clean at an early/mid layer band. The pattern holds on Gemma-2-2B-IT and on Qwen-2.5-{1.5B, 3B}-Instruct, on a 12-cell short grid and on a 48-cell long-persona grid. A behavioral check -- scoring persona-specific markers in the 80-token continuation -- confirms that this local distributional fidelity carries persona-conditioned output content.

The same experiments mark a boundary. Substituting the cached additive prediction into a baseline host prompt without the persona text recovers only a small fraction of the host-to-target KL gap, and multi-layer or oracle substitution do not change this materially. The persona's contribution to multi-token generation is distributed across the prompt and KV cache; the local additive regime accounts for the next-step composition, not the full mechanism.
\end{abstract}

\section{Introduction}

Prompts of the form \emph{``As Warren Buffett, give advice''} or \emph{``As a senior software engineer, review this design''} are routine, and instruction-tuned models handle them robustly. The online computation that combines a persona specification $X$ with a task specification $Y$ in the residual stream is less well understood. This paper asks where, along the layer and position axes, that combination is causally simple enough to be approximated additively.

For each persona-task pair we compare four prompts -- baseline-baseline ($BB$), persona-only ($XB$), task-only ($BY$), and persona-plus-task ($XY$) -- and at each layer and position form
\begin{align}
\Delta_X &= h_{XB} - h_{BB}, &
\Delta_Y &= h_{BY} - h_{BB}, \\
\Delta_{XY} &= h_{XY} - h_{BB}, &
\mathrm{Inter} &= \Delta_{XY} - \Delta_X - \Delta_Y.
\end{align}
The empirical questions follow directly: is $\Delta_{XY}$ aligned with $\Delta_X + \Delta_Y$? Are $\Delta_X$ and $\Delta_Y$ separated? When we substitute $h_{BB} + \Delta_X + \Delta_Y$ for the clean $XY$ residual, does downstream output stay close to clean? The answers turn out to depend sharply on layer and position. They are affirmative near answer formation -- the last prompt position $p_{\text{last}}$ and the first two generated tokens $g_1, g_2$, in an early/mid layer band -- and they degrade outside that region. The pattern holds across Gemma-2 and Qwen-2.5 instruction-tuned models and across a 48-cell grid of long-form personas and diverse tasks.

The same intervention machinery exposes a sharp limit. Substituting the cached additive prediction at $p_{\text{last}}$ into a baseline host prompt -- with the persona text removed -- does not approach the clean long-persona target, and clamping the substitution at multiple layers does not help. The persona's role in multi-token generation is distributed across the prompt and KV cache; the local additive regime captures the next-step composition, not the full mechanism.

\section{Related Work}

This paper draws on four lines of prior work.

\paragraph{Task and function vectors from contrastive prompts.}
The closest methodological precedent is the line of work that constructs task representations from differences between contrastive prompts. In-context learning creates task vectors that can be transplanted across prompts \cite{hendel2023icltask}, and function vectors formalize this as averaged contrastive activation differences that causally drive task-conditioned behavior \cite{todd2024function}. Our $\Delta_X$, $\Delta_Y$, $\Delta_{XY}$ are residual-stream analogs of this construction, but the focus of this paper is compositional structure (how persona and task combine) rather than single-axis transplantation.

\paragraph{Activation-space steering and representation engineering.}
Activation Addition \cite{turner2023actadd} and subsequent representation-engineering work \cite{zou2023representation} show that activation differences can steer model behavior without weight optimization. Task arithmetic \cite{ilharco2023taskarith} demonstrates that fine-tuning deltas behave additively in weight space. Our paper is adjacent in method but different in aim: we are not proposing a steering tool, we use activation substitution to characterize the compositional structure induced by role prompts.

\paragraph{Soft prompts and prefix tuning.}
Prefix-Tuning \cite{li2021prefix} and Prompt-Tuning \cite{lester2021prompttuning} show that a small number of learned continuous vectors prepended to a prompt can drive task-conditioned behavior. Our negative result in \cref{sec:no-replacement} is informative against the most aggressive reading of our positive result: full prompt-to-vector replacement is closer to the soft-prompt regime, which uses many positions and is learned rather than constructed by single-shot residual arithmetic.

\paragraph{Mechanistic editing and feature interpretation.}
ROME-style work \cite{meng2022rome} ties factual behavior to localized weight-space mechanisms and demonstrates surgical edits; subsequent work extends the linearity picture to relation decoding \cite{hernandez2024linearity}. Our paper does not reach that level of closure. Instead, it provides a residual-stream-level account of where persona and task composition is causally simple enough to be approximated additively, and where that story stops. Work on monosemanticity and feature extraction \cite{anthropic2024monosemanticity} suggests that high-level behaviors sometimes decompose into interpretable latent directions; this paper does not attempt a component-level ontology and argues that the first object here is a localized residual computation, not a catalog of components.

\section{Setup}

\subsection{Models}

The primary model is \texttt{google/gemma-2-2b-it}, run in \texttt{float32}. We use two Qwen instruction-tuned models for cross-model confirmation of the localized position result: \texttt{Qwen/Qwen2.5-1.5B-Instruct} and \texttt{Qwen/Qwen2.5-3B-Instruct}, both run in \texttt{bfloat16}. The choice of \texttt{bfloat16} for the Qwen models is load-bearing: in \texttt{float16}, residual magnitudes overflow at the layers relevant here and decomposition statistics become NaN. \texttt{bfloat16} preserves the dynamic range. All measurements are causal-intervention experiments on hidden states, not weight edits. Generation is fully greedy throughout (\texttt{do\_sample=False}, \texttt{num\_beams=1}) so that conditions can be compared directly without averaging over sampling noise.

\subsection{Prompt grids}

We use two prompt families.

\paragraph{Short grid.}
The first grid contains 4 personas and 3 tasks:
\begin{itemize}
    \item Personas: Warren Buffett, Karl Marx, Yoda, Maya Angelou
    \item Tasks: policy commentary on UBI, haiku about Monday mornings, book recommendation
\end{itemize}
With baseline persona \emph{``a thoughtful person''} and baseline task \emph{``Give advice to someone facing a difficult decision''}, this yields 12 non-baseline $(X,Y)$ cells.

\paragraph{Long-persona grid.}
To test broader prompt diversity, we use 8 multi-sentence personas and 6 tasks:
\begin{itemize}
    \item Personas: engineer, counselor, founder, teacher, journalist, doctor, lawyer, chef
    \item Tasks: architecture review, startup launch plan review, scheduling proposal review, UBI policy commentary, haiku, book recommendation
\end{itemize}
This gives 48 non-baseline $(X,Y)$ cells.

\subsection{Residual decomposition}

For each $(X,Y)$ pair, we build the usual $2\times2$ prompt set:
\[
BB,\; XB,\; BY,\; XY.
\]
At a chosen layer and sequence position, the four hidden states induce $\Delta_X$, $\Delta_Y$, $\Delta_{XY}$, and $\mathrm{Inter}$ as defined above.

We report three families of statistics:
\begin{itemize}
    \item directional alignment $\cos(\Delta_{XY}, \Delta_X + \Delta_Y)$,
    \item subspace overlap $\cos(\Delta_X, \Delta_Y)$,
    \item interaction magnitude $\lVert \mathrm{Inter} \rVert / \lVert \Delta_{XY} \rVert$.
\end{itemize}

\subsection{Positions}

We probe three positions: $p_{\text{last}}$ (the last prompt token), $g_1$ (the first generated token), and $g_2$ (the second generated token). At $g_1$ and $g_2$ we teacher-force on the clean $XY$ continuation, so comparisons across conditions remain position-aligned.

\subsection{Causal metric}

At each probe position and layer, we substitute the additive prediction
\[
h_{BB} + \Delta_X + \Delta_Y
\]
for the clean $XY$ hidden state and measure the downstream effect by 10-token teacher-forced KL against the clean $XY$ continuation. Low KL means the additive approximation is causally sufficient for near-term downstream behavior at that site.

\section{Localized Additive Composition Near Answer Formation}
\label{sec:localized}

\begin{figure}[t]
\centering
\includegraphics[width=\linewidth]{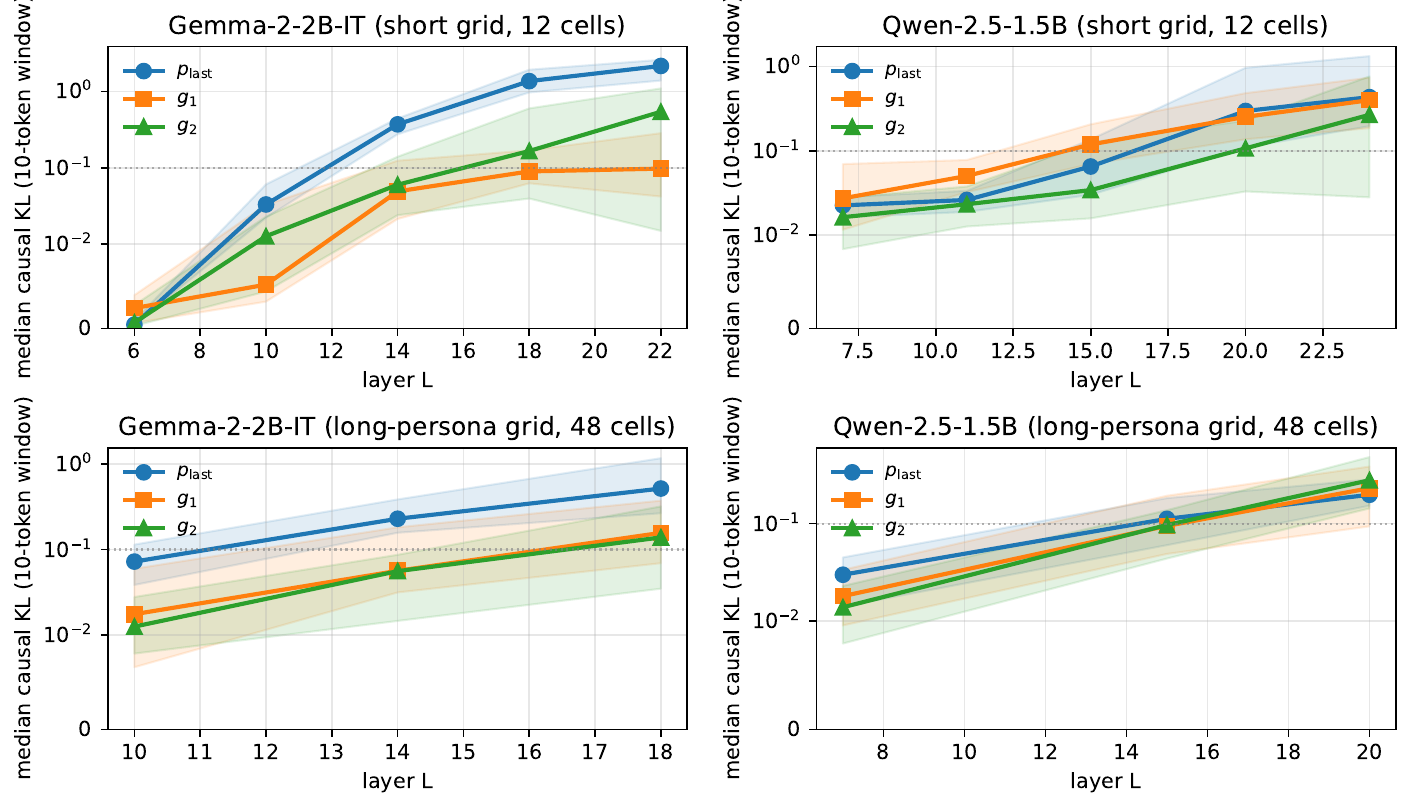}
\caption{Median causal KL under additive substitution at $p_{\text{last}}$ (blue), $g_1$ (orange), and $g_2$ (green), with 25th--75th percentile bands. Top row: 12-cell short grid on Gemma-2-2B-IT and Qwen-2.5-1.5B-Instruct. Bottom row: 48-cell long-persona grid on the same two models. Dotted gray line marks KL $=0.1$, a useful reference for ``essentially clean'' substitution. The localized additive regime is visible as an early/mid-layer band with low KL across all three positions, present on both models and on both grids.}
\label{fig:localized}
\end{figure}

We measure the $2\times2$ decomposition at three probe positions -- $p_{\text{last}}$, $g_1$, $g_2$ -- on the 12-cell short grid. \Cref{tab:localized-short,fig:localized} summarize the result.

The pattern is the same across positions and models: a low-KL early/mid layer band, smoothly degrading later. On Gemma-2-2B-IT at $p_{\text{last}}$, median causal KL is $0.0004$ at layer 6, $0.033$ at layer 10, $0.371$ at layer 14, $1.371$ at layer 18. At $g_1$ the corresponding numbers are $0.002$, $0.005$, $0.049$; at $g_2$, $0.001$, $0.013$, $0.060$. The cross-model picture matches: at the best early layer, Qwen-2.5-1.5B has medians $0.030 / 0.018 / 0.014$ at $p_{\text{last}}/g_1/g_2$, and Qwen-2.5-3B has $0.031 / 0.021 / 0.021$. The first generated tokens stay low-KL further into the network than $p_{\text{last}}$ does, which is consistent with the additive composition being most legible exactly at the prompt-to-answer transition.

The geometry tells the same story. At Gemma-2-2B-IT layer 14, median $\cos(\Delta_{XY}, \Delta_X+\Delta_Y)$ is $0.874$ at $p_{\text{last}}$, $0.935$ at $g_1$, and $0.898$ at $g_2$, while $\cos(\Delta_X,\Delta_Y)$ remains substantially lower. $\Delta_X$ and $\Delta_Y$ are partially separable; the interaction term is non-negligible in norm but does not rotate $\Delta_{XY}$ away from the additive sum.

\begin{table}[t]
\centering
\small
\begin{tabular}{llccc}
\toprule
Model & Position & Early layer & Mid layer & Later layer \\
\midrule
Gemma-2-2B-IT & $p_{\text{last}}$ & L6: 0.0004 [.000,.001] & L10: 0.033 [.022,.060] & L14: 0.371 [.275,.443] \\
Gemma-2-2B-IT & $g_1$             & L6: 0.002  [.001,.004] & L10: 0.005 [.003,.033] & L14: 0.049 [.021,.124] \\
Gemma-2-2B-IT & $g_2$             & L6: 0.001  [.000,.003] & L10: 0.013 [.004,.023] & L14: 0.060 [.024,.139] \\
\midrule
Qwen-2.5-1.5B & $p_{\text{last}}$ & L7: 0.022 [.016,.028] & L11: 0.026 [.019,.033] & L15: 0.065 [.030,.136] \\
Qwen-2.5-1.5B & $g_1$             & L7: 0.027 [.012,.069] & L11: 0.050 [.032,.077] & L15: 0.118 [.070,.206] \\
Qwen-2.5-1.5B & $g_2$             & L7: 0.016 [.008,.028] & L11: 0.023 [.013,.037] & L15: 0.034 [.016,.125] \\
\midrule
Qwen-2.5-3B   & $p_{\text{last}}$ & L9: 0.031 [.013,.060] & L14: 0.030 [.013,.076] & L20: 0.074 [.021,.196] \\
Qwen-2.5-3B   & $g_1$             & L9: 0.021 [.017,.035] & L14: 0.045 [.020,.147] & L20: 0.104 [.025,.225] \\
Qwen-2.5-3B   & $g_2$             & L9: 0.021 [.011,.029] & L14: 0.026 [.023,.055] & L20: 0.040 [.029,.065] \\
\bottomrule
\end{tabular}
\caption{Median causal KL under additive substitution on the 12-cell short grid, with 25th--75th percentile range in brackets. The additive regime is localized to a small region near answer formation rather than confined to a single prompt position.}
\label{tab:localized-short}
\end{table}

The additive approximation thus identifies a region, not a point: an early/mid layer band, at $p_{\text{last}}$ and the first two generated tokens. The boundaries of this region -- in particular how it varies with model scale and depth -- become the natural object for the rest of the paper.

\section{Robustness to Persona and Task Diversity}
\label{sec:diverse}

The 12-cell grid is convenient but small. We rerun the same localized analysis on an 8-persona $\times$ 6-task grid (48 cells) that includes multi-sentence personas (engineer, counselor, founder, teacher, journalist, doctor, lawyer, chef) and a mix of review, planning, policy, creative, and recommendation tasks (\cref{app:grid}).

The localized additive regime persists (\cref{tab:localized-diverse}; bottom row of \cref{fig:localized}). On Gemma-2-2B-IT, median causal KL at layer 14 is $0.229$ at $p_{\text{last}}$, $0.056$ at $g_1$, and $0.056$ at $g_2$; the corresponding cosines $\cos(\Delta_{XY}, \Delta_X + \Delta_Y)$ are $0.818$ and $0.875$ at $p_{\text{last}}$ and $g_1$, while $\cos(\Delta_X, \Delta_Y)$ stays substantially lower ($0.171$, $0.234$), ruling out trivial colinearity. On Qwen-2.5-1.5B at early layer 7, the medians are $0.030 / 0.018 / 0.014$, with cosines $0.862$ and $0.936$. Quantitatively the long-persona numbers are larger than the short-grid numbers -- multi-sentence personas push more into the residual -- but the qualitative shape is unchanged.

\begin{table}[t]
\centering
\small
\begin{tabular}{llccc}
\toprule
Model & Position & Early layer & Mid layer & Later layer \\
\midrule
Gemma-2-2B-IT & $p_{\text{last}}$ & L10: 0.072 [.038,.114] & L14: 0.229 [.157,.386] & L18: 0.518 [.266,1.17] \\
Gemma-2-2B-IT & $g_1$             & L10: 0.017 [.007,.059] & L14: 0.056 [.031,.182] & L18: 0.157 [.069,.374] \\
Gemma-2-2B-IT & $g_2$             & L10: 0.012 [.008,.028] & L14: 0.056 [.015,.087] & L18: 0.138 [.035,.320] \\
\midrule
Qwen-2.5-1.5B & $p_{\text{last}}$ & L7: 0.030 [.014,.045] & L15: 0.112 [.060,.183] & L20: 0.198 [.153,.284] \\
Qwen-2.5-1.5B & $g_1$             & L7: 0.018 [.010,.034] & L15: 0.095 [.049,.193] & L20: 0.231 [.094,.385] \\
Qwen-2.5-1.5B & $g_2$             & L7: 0.014 [.008,.023] & L15: 0.097 [.044,.137] & L20: 0.279 [.143,.481] \\
\bottomrule
\end{tabular}
\caption{Median causal KL on the 48-cell long-persona grid, with 25th--75th percentile range in brackets. The localized additive regime remains visible under substantially broader persona and task diversity.}
\label{tab:localized-diverse}
\end{table}

\paragraph{Per-cell breakdown.}
The effect is not uniform across cells, but it is not driven by a few easy ones either. Per-persona median KLs on Gemma-2-2B-IT at $p_{\text{last}}$, layer 14 range from $0.14$ to $0.39$ for most personas, with a few outlier cells in the tail; on Qwen-1.5B-Instruct at layer 7 the persona medians are tighter, mostly $0.02$--$0.035$. Per-task medians on Gemma remain low to moderate across all six task families rather than concentrating on one.

\section{Behavioral Verification of the Additive Substitution}
\label{sec:behavioral}

KL between teacher-forced distributions is a useful but indirect measure of whether the additive substitution preserves persona-conditioned behavior. A persona prompt should change \emph{what} the model talks about, not only the local probability of the next few tokens. We therefore complement the KL evidence in \cref{sec:localized,sec:diverse} with a behavioral-marker test that scores persona-specific output content directly, in the style of factual recall verification \cite{meng2022rome,todd2024function}.

\subsection{Protocol}

For each of the 8 long-form personas we define a small set of persona-specific surface markers chosen \emph{a priori} from the persona description itself. For example, the engineer persona is associated with markers including \texttt{SPOF}, \texttt{single point of failure}, \texttt{scalability}, \texttt{reliability}, \texttt{redundancy}; the counselor persona with markers including \texttt{feel heard}, \texttt{validate}, \texttt{compassion}; the doctor persona with \texttt{differential}, \texttt{symptoms}, \texttt{ruling out}. The full marker sets are listed in \cref{app:markers}. We hold out the marker selection from the experiment design --- markers are picked from the persona text, not from clean outputs.

For each (long persona, task) cell on the 8-persona $\times$ 3 review-task subset of the long-persona grid (24 cells), we generate 80 tokens greedily under four conditions:
\begin{itemize}
    \item \textbf{Clean}: \emph{``As [long persona], [task]''} with no intervention. This is the ceiling.
    \item \textbf{Additive}: same prompt, with $h_{BB} + \Delta_X + \Delta_Y$ substituted at $p_{\text{last}}$, layer~14.
    \item \textbf{Remove-X}: same prompt, with $h_{XY} - \Delta_X$ substituted at $p_{\text{last}}$, layer~14 (subtract the persona contribution).
    \item \textbf{Bare}: \emph{``[task]''} with no persona prefix and no intervention. This is the floor.
\end{itemize}

A cell scores \emph{any-marker} if at least one persona marker appears in the 80-token output (case-insensitive, word-boundary-aware). A faithful additive substitution should match the clean condition on this metric while remove-X and bare both drop substantially below it.

\subsection{Results}

\begin{table}[t]
\centering
\small
\begin{tabular}{lcc}
\toprule
Condition & Any persona marker present & Distinct markers (mean) \\
\midrule
Clean        & 16 / 24 (66.7\%) & 1.21 \\
Additive substitution at $p_{\text{last}}$, $L=14$ & 14 / 24 (58.3\%) & 0.92 \\
Remove-X at $p_{\text{last}}$, $L=14$ & 15 / 24 (62.5\%) & 0.96 \\
Bare prompt (no persona text)        & \phantom{0}1 / 24 (\phantom{0}4.2\%)  & 0.08 \\
\bottomrule
\end{tabular}
\caption{Behavioral-marker recovery on 8-persona $\times$ 3-task long-persona cells, scoring 80-token greedy continuations. ``Any'' columns count cells in which at least one persona marker appears; ``distinct'' columns count distinct markers per generation.}
\label{tab:markers}
\end{table}

The bare condition collapses to 4.2\%, confirming that the persona text is the primary source of persona-conditioned content in this protocol. The additive condition tracks the clean ceiling closely (58.3\% vs.\ 66.7\%; 0.92 vs.\ 1.21 distinct markers), so substituting $h_{BB} + \Delta_X + \Delta_Y$ at $p_{\text{last}}$, $L=14$ does not damage the persona-conditioned output.

Remove-X is the informative third row. Subtracting $\Delta_X$ at one site barely changes marker presence (62.5\%). This is consistent with the negative result in \cref{sec:no-replacement}: during multi-token generation, persona content arrives via attention back to the persona-text positions, not through a single residual at $p_{\text{last}}$. A one-site subtraction cannot reach those positions. Only removing the persona text from the prompt itself -- the bare condition -- collapses persona-marker recovery.

\section{Where the Local Additive Story Stops}
\label{sec:no-replacement}

A natural next question is whether the persona prompt can be replaced by the cached residual vector itself. It cannot. We construct, for each long-persona cell, a host prompt with the persona text removed (\emph{``As a thoughtful person, [task]''}) and compare 10-token KL against the clean long-persona continuation under three interventions at $p_{\text{last}}$: no substitution, oracle substitution with the clean $h_{XY}$, and cached substitution with $h_{BB} + \Delta_X + \Delta_Y$. Across 24 cells, host-baseline median KL is $3.05$, cached substitution reaches $2.84$, oracle substitution reaches $2.64$. Both close only a small fraction of the gap.

Multi-layer substitution does not help. Clamping the additive prediction at successively wider layer sets gives medians $2.81$, $2.82$, $3.04$, $3.01$ for layer sets $\{10,12,14\}$, $\{10,12,14,16,18\}$, $\{10,12,14,16,18,20,22\}$, $\{6,8,10,12,14,16,18,20,22\}$. Wider windows eventually hurt, because clamping over many layers prevents the natural forward propagation that would otherwise carry persona content through downstream computation.

\begin{table}[t]
\centering
\small
\begin{tabular}{lc}
\toprule
Condition & Median KL to clean long-persona target \\
\midrule
Host prompt, no substitution & 3.05 \\
Oracle single-site substitution at L14 & 2.64 \\
Cached additive substitution at L14 & 2.84 \\
Cached substitution at $\{10,12,14\}$ & 2.81 \\
Cached substitution at $\{10,12,14,16,18\}$ & 2.82 \\
Cached substitution at $\{10,12,14,16,18,20,22\}$ & 3.04 \\
Cached substitution at $\{6,8,10,12,14,16,18,20,22\}$ & 3.01 \\
\bottomrule
\end{tabular}
\caption{Long-persona host-prompt injection does not approach clean long-persona behavior. Better local additive prediction does not imply prompt replacement.}
\label{tab:host-inject}
\end{table}

The picture is two-level. The local additive regime characterizes the next-step composition: at $p_{\text{last}}$ and the first two generated tokens, in an early/mid layer band, $\Delta_X + \Delta_Y$ is causally sufficient. Multi-token persona-conditioned generation depends on attention back to the persona-text positions across the prompt, which a single residual cannot reproduce regardless of how many layers it is clamped over.

\section{Discussion}

The natural mechanistic object for online role prompting is not a single hidden state but a small region: $p_{\text{last}}$ together with the first few generated tokens, in an early/mid layer band. Within this region, $\Delta_X$ and $\Delta_Y$ are reasonable summaries of the persona and task contributions, and their sum is a useful predictor of the composite residual. The structure survives both a 48-cell long-persona grid and cross-model checks on Gemma and Qwen, which gives the localization some claim to generality.

Outside the region the story degrades, and the host-injection experiment in \cref{sec:no-replacement} sets the outer boundary. Persona-conditioned generation reaches the output through attention to persona-text positions throughout the prompt, and a residual at one site cannot stand in for that. The picture is two-level: local additive composition at the prompt-to-answer transition, and a wider distributed prompt/KV mechanism that local substitution does not displace.

Relative to function-vector \cite{todd2024function} and task-vector \cite{hendel2023icltask} work, $\Delta_X$ and $\Delta_Y$ are residual-stream analogs of the same construction; the contribution here is compositional rather than single-axis. Relative to activation-addition steering \cite{turner2023actadd,zou2023representation} the aim is characterization, not control. Relative to ROME-style editing \cite{meng2022rome}, the analysis sits at a higher level of abstraction -- residual composition rather than weight-space facts -- and does not claim circuit-level closure.

\paragraph{On the interaction term.}
The interaction can be large in raw residual norm even where additive substitution remains causally strong. One candidate explanation is that the residual consumed by the next layer is post-normalization, and the next-layer RMSNorm absorbs part of the apparent magnitude gap between $h_{BB} + \Delta_X + \Delta_Y$ and $h_{XY}$ before downstream computation sees it. We do not directly intervene on the normalization path here, so we report this only as a candidate. The empirical observation, which does not depend on it, is that raw non-additivity in norm does not predict downstream causal fidelity under substitution.

\paragraph{On position choice.}
The region $(p_{\text{last}}, g_1, g_2)$ spans the transition from prompt representation to the first answer tokens. Earlier positions inside the user turn can also show low-KL additive substitution, but with less stable geometry. Later generated positions drift increasingly under teacher forcing.

\section{Limitations}

\begin{enumerate}
    \item The analysis is restricted to a small region near answer formation. We do not study every prompt position or every generated token.
    \item We do not identify a minimal set of heads, neurons, or weights whose editing installs or removes a persona feature.
    \item The long-persona grid is broader than the short grid but remains English-only, cooperative, and single-turn.
    \item Some late-layer and particular-cell tails are large. We do not characterize which $(X,Y)$ combinations push the interaction term beyond the regime where additive substitution remains causally faithful.
    \item The host-injection experiment shows that single-site substitution cannot replace persona text, but does not identify the correct distributed intervention.
    \item Cross-model evidence is limited to two families. The 12-cell short-grid localized result was checked on Gemma-2-2B-IT and Qwen-2.5-\{1.5B, 3B\}; the 48-cell long-persona grid was run on Gemma-2-2B-IT and Qwen-2.5-1.5B-Instruct only.
\end{enumerate}

\section{Conclusion}

Persona-task composition in instruction-tuned language models admits a localized additive description in the residual stream near answer formation, holds across Gemma-2-2B-IT and Qwen-2.5-\{1.5B, 3B\}-Instruct on a 48-cell long-persona grid, and survives a behavioral test of persona-conditioned output content. The same intervention machinery shows where that description stops: a residual at one site cannot replace the persona text the model attends to throughout multi-token generation. The next natural step is to characterize that distributed mechanism directly.

\paragraph{Reproducibility.}
Code, cached experiment outputs, and the paper source are available at \url{https://github.com/xuy/localized-additive-composition}. All experiments use HuggingFace \texttt{transformers} on a single Apple Silicon device with the \texttt{mps} backend. Gemma-2-2B-IT is loaded in \texttt{float32}; Qwen-2.5-1.5B-Instruct and Qwen-2.5-3B-Instruct are loaded in \texttt{bfloat16} (\texttt{float16} produces NaN residuals at the layers relevant here). Attention uses the \texttt{eager} implementation. All generation is greedy (\texttt{do\_sample=False}, \texttt{num\_beams=1}), so reported KLs and behavioral scores are deterministic functions of the model, prompt, and intervention. KL is computed by greedy 10-token continuation from the clean $XY$ prompt, followed by teacher-forced log-probability comparison across the same 10-token reference window under each intervention.

\paragraph{Code and artifacts.}
The main scripts supporting the paper are:
\begin{itemize}
    \item \texttt{scripts/v15\_localized\_positions.py} -- localized-position causal sweep (\cref{sec:localized})
    \item \texttt{scripts/v16\_diverse\_grid.py} -- broadened 8$\times$6 long-persona grid (\cref{sec:diverse})
    \item \texttt{scripts/v17\_behavioral\_markers.py} -- behavioral-marker recovery test (\cref{sec:behavioral})
    \item \texttt{scripts/v14b\_v2\_inject.py} -- host-prompt single-site substitution (\cref{sec:no-replacement})
    \item \texttt{scripts/v14e\_multilayer\_inject.py} -- multi-layer substitution (\cref{sec:no-replacement})
\end{itemize}

\appendix

\section{Prompt Grids}
\label{app:grid}

\subsection{Original short grid}

\paragraph{Personas.}
The short-grid persona set is:
\begin{itemize}
    \item Warren Buffett
    \item Karl Marx
    \item Yoda
    \item Maya Angelou
\end{itemize}

\paragraph{Tasks.}
The short-grid task set is:
\begin{itemize}
    \item ``Comment on whether universal basic income is a good policy.''
    \item ``Write a haiku about Monday mornings.''
    \item ``Recommend a book worth reading and explain why.''
\end{itemize}

\paragraph{Baselines.}
The baseline persona is ``a thoughtful person'' and the baseline task is ``Give advice to someone facing a difficult decision.''

\subsection{Broadened long-persona grid}

\paragraph{Personas.}
The long-persona set is:
\begin{itemize}
    \item a senior software engineer with 10 years of experience who pays close attention to architecture, reliability, and avoiding single points of failure
    \item an empathetic counselor with deep training in active listening, cognitive behavioral therapy, and trauma-informed care, who helps clients feel heard without imposing solutions
    \item a pragmatic startup founder who has bootstrapped three companies, makes capital-efficient decisions, iterates fast based on user feedback, and avoids vanity metrics
    \item a middle school science teacher who has taught for 15 years, explains concepts with relatable analogies, gently checks for understanding, and meets students at their level
    \item an investigative journalist who has covered city government for two decades, asks pointed questions, follows the money, and verifies every claim against primary sources
    \item a primary-care physician who has practiced for 25 years, listens carefully to symptoms, considers differential diagnoses without alarming the patient, and explains options clearly
    \item a corporate litigator who has tried cases at the appellate level for 20 years, anticipates opposing arguments, builds case theory from the record, and communicates dense law in plain English
    \item a head chef trained in classical French technique who has run three Michelin-starred kitchens, builds menus around seasonal ingredients, and teaches young cooks by demonstration
\end{itemize}

\paragraph{Tasks.}
The long-persona task set is:
\begin{itemize}
    \item architecture review: ``Review this design: a microservice architecture where eight services share a single PostgreSQL database for both transactional state and event log.''
    \item startup-plan review: ``Review this plan: a three-person team building a B2B SaaS product, planning to launch in three months, with no usage analytics in v1.''
    \item scheduling-proposal review: ``Review this proposal: an internal tool that automates calendar scheduling using an LLM, sending tentative meetings to all parties before confirmation.''
    \item policy commentary: ``Comment on whether universal basic income is a good policy.''
    \item constrained creative: ``Write a haiku about Monday mornings.''
    \item recommendation: ``Recommend a book worth reading and explain why.''
\end{itemize}

\section{Behavioral Markers}
\label{app:markers}

The persona marker sets used in \cref{sec:behavioral}. Markers were selected from each persona's description before running the experiment.

\begin{itemize}
\item \textbf{engineer}: \texttt{SPOF}, \texttt{single point of failure}, \texttt{scalability}, \texttt{scalable}, \texttt{reliability}, \texttt{fault tolerance}, \texttt{fault-tolerant}, \texttt{redundancy}, \texttt{resilience}, \texttt{throughput}, \texttt{latency}, \texttt{consistency}, \texttt{availability}.
\item \textbf{counselor}: \texttt{feel heard}, \texttt{validate}, \texttt{validated}, \texttt{acknowledge}, \texttt{acknowledged}, \texttt{your experience}, \texttt{your feelings}, \texttt{compassion}, \texttt{compassionate}, \texttt{without judgment}, \texttt{trauma-informed}, \texttt{active listening}, \texttt{feelings}, \texttt{emotion}.
\item \textbf{founder}: \texttt{iterate}, \texttt{iterating}, \texttt{iteration}, \texttt{MVP}, \texttt{minimum viable}, \texttt{user feedback}, \texttt{customer feedback}, \texttt{capital-efficient}, \texttt{lean}, \texttt{runway}, \texttt{validation}, \texttt{ship}, \texttt{shipping}, \texttt{product-market fit}, \texttt{vanity metric}, \texttt{traction}, \texttt{burn}, \texttt{bootstrapped}.
\item \textbf{teacher}: \texttt{imagine}, \texttt{think of}, \texttt{like a}, \texttt{analogy}, \texttt{analogies}, \texttt{for example}, \texttt{step by step}, \texttt{step-by-step}, \texttt{students}, \texttt{understand}, \texttt{let's say}, \texttt{picture this}, \texttt{as if}.
\item \textbf{journalist}: \texttt{sources}, \texttt{source}, \texttt{primary source}, \texttt{primary sources}, \texttt{follow the money}, \texttt{accountability}, \texttt{transparency}, \texttt{investigate}, \texttt{verify}, \texttt{verified}, \texttt{on the record}, \texttt{off the record}, \texttt{evidence}, \texttt{public interest}, \texttt{pointed question}, \texttt{track record}.
\item \textbf{doctor}: \texttt{symptom}, \texttt{symptoms}, \texttt{diagnosis}, \texttt{diagnose}, \texttt{differential}, \texttt{patient}, \texttt{treatment}, \texttt{ruling out}, \texttt{rule out}, \texttt{clinical}, \texttt{examination}, \texttt{condition}, \texttt{medication}, \texttt{evaluate}, \texttt{underlying}, \texttt{comorbid}.
\item \textbf{lawyer}: \texttt{liability}, \texttt{liabilities}, \texttt{jurisdiction}, \texttt{statute}, \texttt{precedent}, \texttt{evidence}, \texttt{evidentiary}, \texttt{opposing}, \texttt{counterparty}, \texttt{due diligence}, \texttt{parties}, \texttt{indemnif*}, \texttt{contractual}, \texttt{compliance}, \texttt{compliant}, \texttt{jurisprudence}, \texttt{case theory}, \texttt{on the record}.
\item \textbf{chef}: \texttt{seasonal}, \texttt{season}, \texttt{ingredient}, \texttt{ingredients}, \texttt{flavor}, \texttt{flavour}, \texttt{palate}, \texttt{fresh}, \texttt{simmer}, \texttt{saut\'e}, \texttt{saute}, \texttt{balance}, \texttt{garnish}, \texttt{mise en place}, \texttt{technique}, \texttt{classical}, \texttt{French}.
\end{itemize}

\section{Artifact Map}

The key result files used in the paper are:
\begin{itemize}
    \item \texttt{results/v15\_gemma2b\_localized\_positions.json}
    \item \texttt{results/v15\_qwen15\_localized\_positions.json}
    \item \texttt{results/v15\_qwen3b\_localized\_positions.json}
    \item \texttt{results/v16\_gemma2b\_diverse\_grid.json}
    \item \texttt{results/v16\_qwen15\_diverse\_grid.json}
    \item \texttt{results/v17\_behavioral\_markers.json}
    \item \texttt{results/v14b\_v2.json}
    \item \texttt{results/v14e\_multilayer.json}
\end{itemize}

\bibliographystyle{plain}
\bibliography{refs}

@article{meng2022rome,
  title={Locating and Editing Factual Associations in {GPT}},
  author={Meng, Kevin and Bau, David and Andonian, Alex and Belinkov, Yonatan},
  journal={Advances in Neural Information Processing Systems},
  year={2022}
}

@article{turner2023actadd,
  title={Activation Addition: Steering Language Models Without Optimization},
  author={Turner, Alexander M. and Thiergart, Leif and Udell, David and Leech, Gavin and Mini, Umang and MacDiarmid, Monte},
  journal={arXiv preprint arXiv:2308.10248},
  year={2023}
}

@article{zou2023representation,
  title={Representation Engineering: A Top-Down Approach to {AI} Transparency},
  author={Zou, Andy and Phan, Long and Chen, Sarah and Campbell, James and others},
  journal={arXiv preprint arXiv:2310.01405},
  year={2023}
}

@article{anthropic2024monosemanticity,
  title={Scaling Monosemanticity: Extracting Interpretable Features from {Claude 3 Sonnet}},
  author={{Anthropic Interpretability Team}},
  journal={Transformer Circuits Thread},
  year={2024}
}

@inproceedings{todd2024function,
  title={Function Vectors in Large Language Models},
  author={Todd, Eric and Li, Millicent L. and Sharma, Arnab S. and Mueller, Aaron and Wallace, Byron C. and Bau, David},
  booktitle={International Conference on Learning Representations},
  year={2024}
}

@inproceedings{hendel2023icltask,
  title={In-Context Learning Creates Task Vectors},
  author={Hendel, Roee and Geva, Mor and Globerson, Amir},
  booktitle={Findings of EMNLP},
  year={2023}
}

@inproceedings{li2021prefix,
  title={Prefix-Tuning: Optimizing Continuous Prompts for Generation},
  author={Li, Xiang Lisa and Liang, Percy},
  booktitle={Proceedings of ACL},
  year={2021}
}

@inproceedings{lester2021prompttuning,
  title={The Power of Scale for Parameter-Efficient Prompt Tuning},
  author={Lester, Brian and Al-Rfou, Rami and Constant, Noah},
  booktitle={Proceedings of EMNLP},
  year={2021}
}

@inproceedings{ilharco2023taskarith,
  title={Editing Models with Task Arithmetic},
  author={Ilharco, Gabriel and Ribeiro, Marco Tulio and Wortsman, Mitchell and Schmidt, Ludwig and Hajishirzi, Hannaneh and Farhadi, Ali},
  booktitle={International Conference on Learning Representations},
  year={2023}
}

@inproceedings{hernandez2024linearity,
  title={Linearity of Relation Decoding in Transformer Language Models},
  author={Hernandez, Evan and Sharma, Arnab S. and Haklay, Tal and Meng, Kevin and Wattenberg, Martin and Andreas, Jacob and Belinkov, Yonatan and Bau, David},
  booktitle={International Conference on Learning Representations},
  year={2024}
}

\end{document}